\documentclass[runningheads]{llncs}

\usepackage{eccv}

\usepackage{eccvabbrv}

\usepackage{graphicx}
\usepackage{booktabs}

\usepackage{colortbl}
\usepackage{color}
\usepackage[accsupp]{axessibility}  

\def\name{Cascade-Zero123\xspace}
\def\stageonename{Base-0123\xspace}
\def\stagetwoname{Refiner-0123\xspace}

\definecolor{Lightblue}{RGB}{218,232,252}

\usepackage{hyperref}

\usepackage{orcidlink}
\usepackage{marvosym}
\usepackage{ifsym}
\renewcommand\footnotemark{}

\begin{document}

\title{Cascade-Zero123: One Image to Highly Consistent 3D with Self-Prompted Nearby Views} 

\titlerunning{Cascade-Zero123}

\author{Yabo Chen\inst{1}\textsuperscript{*} \and Jiemin Fang\inst{2}\textsuperscript{*$\dagger$} \and Yuyang Huang\inst{1} \and Taoran Yi\inst{3} \and
\\Xiaopeng Zhang\inst{2}\textsuperscript{\Letter} \and Lingxi Xie\inst{2} \and Xinggang Wang\inst{3} \and Wenrui Dai\inst{1}\textsuperscript{\Letter} \and \\Hongkai Xiong\inst{1} \and Qi Tian\inst{2}\thanks{\textsuperscript{*}Equal contribution. \textsuperscript{$\dagger$}Project lead.\\
\textsuperscript{\Letter}Correspondence to Xiaopeng Zhang and Wenrui Dai.}}
\authorrunning{Y. Chen et al.}
\institute{Shanghai Jiao Tong University, Shanghai, China \and Huawei Inc., Shenzhen, China \and Huazhong University of Science and Technology, Wuhan, China\\
{\tt\small\{chenyabo, huangyuyang, daiwenrui, xionghongkai\}@sjtu.edu.cn}\\
{\tt\small{jaminfong@gmail.com}}, {\tt\small\{taoranyi, xgwang\}@hust.edu.cn}\\{\tt\small{zxphistory@gmail.com}}, {\tt\small{198808xc@gmail.com}}, {\tt\small{tian.qi1@huawei.com}\\}
}

\maketitle
\begin{center}
\centering
\includegraphics[width=1.0\linewidth]{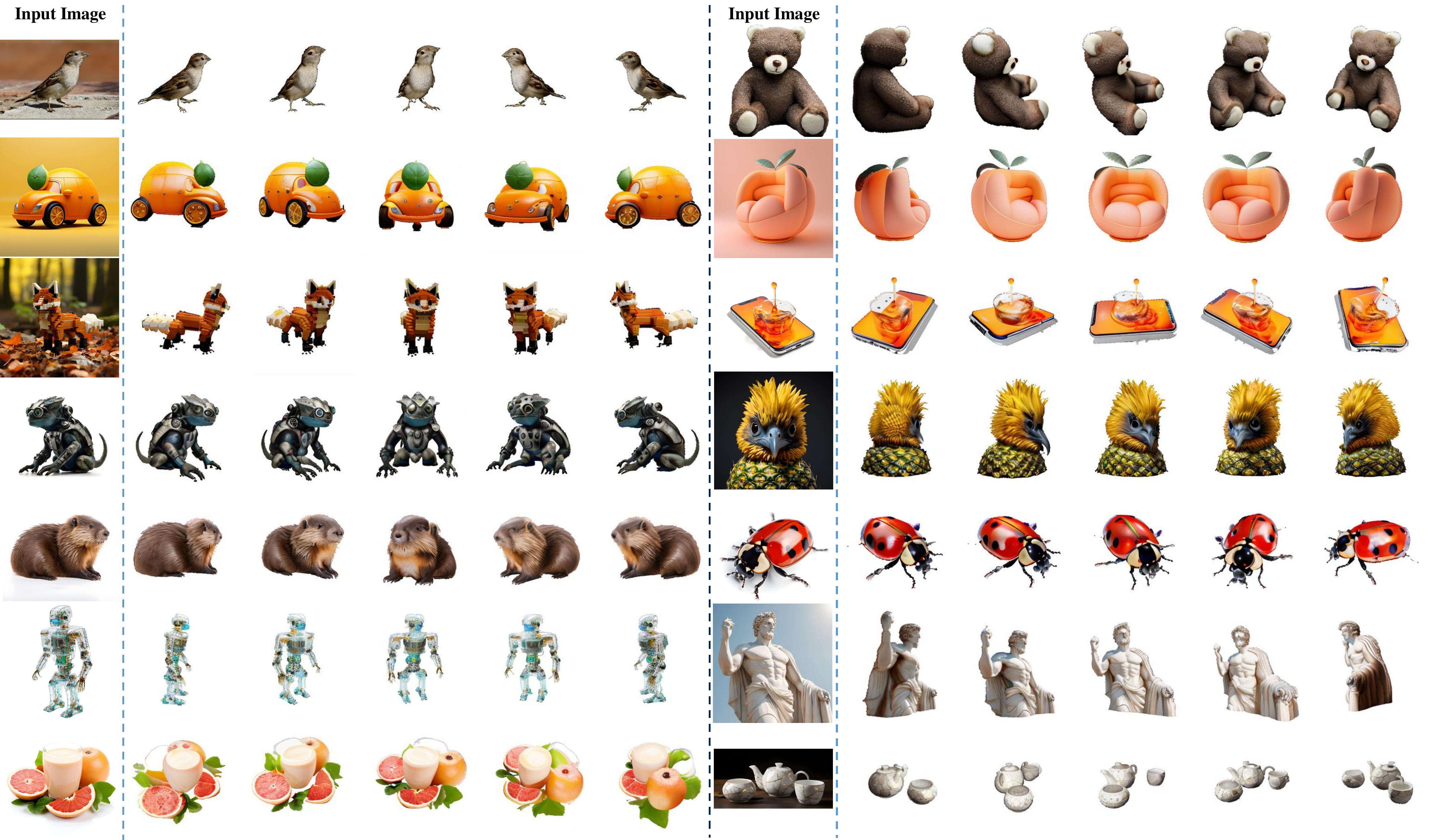}
\captionof{figure}{Rather than adopting limited input information, which Zero-1-to-3~\cite{liu2023zero1to3} generation pipeline only has a single-view source image, \name progressively extracts the 3D information from more condition images by self-prompting. View-consistent images can be generated in a cascade manner. \name shows the strong capability on various complex objects, \eg insects, robots, or multiple objects stacked. 
}
\label{fig: fig1_consist}
\end{center}

\begin{abstract}
Synthesizing multi-view 3D from one single image is a significant but challenging task. Zero-1-to-3 methods have achieved great success by lifting a 2D latent diffusion model to the 3D scope. The target-view image is generated with a single-view source image and the camera pose as condition information. 
However, due to the high sparsity of the single input image, Zero-1-to-3 tends to produce geometry and appearance inconsistency across views, especially for complex objects. To tackle this issue, we propose to supply more condition information for the generation model but in a self-prompt way. A cascade framework is constructed with two Zero-1-to-3 models, named \name, which progressively extract 3D information from the source image. Specifically, several nearby views are first generated by the first model and then fed into the second-stage model along with the source image as generation conditions. With amplified self-prompted condition images, our \name generates more consistent novel-view images than Zero-1-to-3. Experiment results demonstrate remarkable promotion, especially for various complex and challenging scenes, involving insects, humans, transparent objects, and stacked multiple objects \etc. More demos and code are available at \url{https://cascadezero123.github.io}.
\end{abstract}

\section{Introduction}
\label{sec:intro}
Generating the 3D object from a single image has become an appealing research topic, for its flexibility and convenience of creating 3D assets, which can be applied to various real-world applications including virtual reality, computer gaming, movie and animation production~\cite{Paliwal2023ReShader, roessle2023ganerf,zheng2024mvd2,melaskyriazi2024im3d,vainer2024collaborative,kant2024spad,spiegl2024viewfusion,tang2024lgm,chen20242l3,chen2024morphable,xu2024agg,liu2024carstudio,li2024learning,pan2024gd2nerf}. However, due to the input information is too limited -- only one image, it is quite challenging to reconstruct a high-fidelity 3D object. Many research works attempt to lift strong generation power from 2D large latent diffusion models~\cite{rombach2022stablediffusion, nichol2021glide,saharia2022photorealistic,ramesh2022hierarchical} to the 3D fields~\cite{lin2023magic3d,wang2023score,chen2023fantasia3d,wang2023prolificdreamer,xu2023dream3d,shi2023mvdream,zhao2023efficientdreamer,armandpour2023re,jun2023shap,nichol2022point,gupta3dgen,gao2022get3d,luo2023scalable,liu2023deceptivenerf,wu20234dgaussians,yang2023deformable3dgs,GaussianDreamerPro,yang2024gaussianobject,GaussianEditor,wu2024blockfusion,pan2024fast,alldieck2024score,wu2023ifusion,zeng2023paint3d}, bringing possibilities to untractable 3D generation tasks.

\begin{figure}[t]
\begin{center}
\includegraphics[width=\textwidth]
{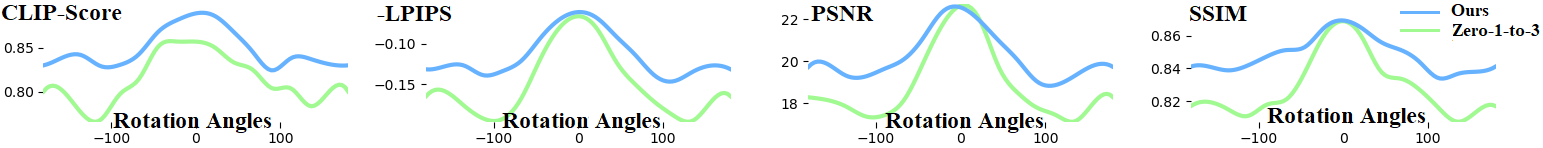}
\end{center}
\caption{The performance comparison of Zero-1-to-3~\cite{liu2023zero1to3} and our methods on Google Scanned Object~\cite{downs2022gso} with different camera pose rotation angles. 
When the camera pose changes drastically, the synthesis quality of Zero-1-to-3 will drop drastically. But our method can promote the synthesis quality in all the transition ranges of camera poses.
}
\label{fig2_rotation_gso}
\end{figure}

As a representative work, Zero-1-to-3~\cite{liu2023zero1to3} innovatively proposes to generate novel view images by conditioning the diffusion model with the source image and corresponding camera pose transition. The Zero-1-to-3 diffusion model is tuned on a large-scale multi-view 3D dataset, Objaverse~\cite{deitke2023objaverse,objaverseXL}, and has achieved remarkable success and generalizability in synthesizing novel views from any single image. However, still suffering from the information sparsity of a single image, Zero-1-to-3 may inevitably fail to keep geometry or appearance consistency across different views, especially for objects with complex structures. As shown in Fig.~\ref{fig2_rotation_gso}, when the camera pose changes drastically, \eg rotating more than $45^\circ$, it is hard to guarantee a high synthesis quality.

Providing more input information beyond a single image will undoubtedly ease the difficulty of generating 3D objects. We expect to improve the Zero-1-to-3 model by supplying more condition images from different views but in a self-prompting way. The main idea is to design a cascade framework to achieve this goal, which contains two Zero-1-to-3 models, \ie \stageonename and \stagetwoname. The \stageonename model is conditioned to generate multiple images from various views. These images are constrained in a small range, \eg $45^\circ$ in our implementation, to guarantee reliable quality as observed in Fig.~\ref{fig2_rotation_gso}. The \stagetwoname model takes in all these generated images along with the source image to output the target-view image. We name the proposed framework as \name. Rather than directly generating the target view as in Zero-1-to-3, \name lowers the synthesis difficulty by progressively digging out the 3D information starting from the input source image. To further enhance the \stageonename model, we propose to update its parameters with a self-distillation manner from \stagetwoname.
As shown in Fig.~\ref{fig2_rotation_gso} and Fig.~\ref{fig: fig1_consist}, our method can consistently promote the synthesis quality in all the transition ranges of camera poses. 

Our contributions can be summarized as follows:

\begin{itemize}
\item We propose a cascade framework to progressively extract 3D information from a single image, preventing inconsistency caused by drastic view transitions.

\item We design a self-prompting method that generates multiple nearby viewpoints, providing reliable condition information for the second-stage \stagetwoname model. 

\item Our proposed \name has shown notable promotion on complex scenes that are challenging for Zero-1-to-3, such as insects, humans, robots, or multiple objects stacked together.
\end{itemize}

\label{sec:Methods}

\section{Related Work}
\label{sec:relatedworks}
\subsection{Single Image to 3D}
Many researchers have studied the tasks of generating 3D models and achieving novel view synthesis using only a single image~\cite{woo2023harmonyview,simon2024hypervoltran,zhang2023repaint123,kocsis2023intrinsic,li2023valid,wu2023hyperdreamer,liu2023one2345,yu2023boosting3d,chen2023singleview,shi2023tosshighquality} and text~\cite{sanghi2022clip,jain2022zero,michel2022text2mesh,lei2022tango,mohammad2022clip,wang2022clip,chen2023text,tang2023dreamgaussian,poole2022dreamfusion,GaussianDreamer,yu2024boostdream,chen2024sketch2nerf}. Some researchers directly train a 3D model on 3D data~\cite{chan2023genvs,xiang20233daware,lee2023localaware,nichol2022point,jun2023shapee,qian2024pushing,huang2024zeroshape}, but they tend to have good generation quality only on scenes similar to the training set.

Recently, by constructing a conditional latent diffusion model based on camera viewpoints, many works have made it possible to pre-train a single image-to-3D model~\cite{liu2023zero1to3,shen2023anything,tang2023make,xu2023neurallift,liu2023one,melas2023realfusion,qian2023magic123,hamdi2023sparf,shi2023zero123++,lin2023consistent123,shi2023toss,sargent2023zeronvs,liu2023syncdreamer,long2023wonder3d,ye2023consistent,weng2023zeroavatar,hu2023humanliff,yang2023consistnet,weng2023consistent123,tang2024mvdiffusion}. Zero-1-to-3~\cite{liu2023zero1to3} learns from large-scale multi-view images~\cite{deitke2023objaverse} to build the geometric priors of large-scale diffusion models. Zero-1-to-3 can lift various images that training sets have never been seen before to 3D with good quality. 
After that, many works have utilized Zero-1-to-3 as a module to enhance the quality of meshes or 3D models. Magic123~\cite{qian2023magic123} combines the capabilities of Zero-1-to-3 and stable diffusion together to generate 3D models. One-2-3-45~\cite{liu2023one} also leverages Zero-1-to-3 to generate different views to assist in mesh generation. However, these methods simply use it as pre-trained 3D diffusion model tools with fixed checkpoints.

There have also been efforts to improve Zero-1-to-3. Approaches like Consistent1-to-3~\cite{ye2023consistent}, aim to enhance the consistency of view generation by introducing priors during the denoising process in view-conditioned diffusion models. However, they add models like additional Transformers to render new views which will result in poor generalization due to these additional model's capability, they may not integrate well with generalized latent diffusion models. In addition, the training of additional models usually costs a lot.

Therefore, we argue that Zero-1-to-3 itself has the ability to provide 3D priors and establish 3D consistency, albeit requiring a progressive manner to achieve it. By using self-prompted condition images, Zero-1-to-3 can generate images of higher quality and higher consistency while also maintaining its strong open-set generalization capabilities brought by the large-scale latent diffusion model.

\subsection{Multi-stage Diffusion Models}
Cascade networks have been widely used in boosting the performance of models in the computer vision field, such as cascade RCNN~\cite{Cai2019cascadercnn}, cascade DETR~\cite{ye2023cascade}, and so on~\cite{li2021poserecong,li2021hierachi}. After large-scale latent diffusion models become popular, many influential works also use cascade networks to improve the generation quality of models. For example, DeepFloyd-IF~\cite{Saharia2022deepfloydif} achieves high-resolution and high-detail image generation by constructing three cascade pixel diffusion modules: a base model that generates a 64$\times$64 pixel image based on text prompts, and two super-resolution models. Furthermore, SDXL~\cite{Podell2023sdxl} uses a cascade structure, in which a base model aligned with a refiner in the latent space to generate higher-quality and higher-resolution images. I2VGen-XL~\cite{2023i2vgenxl} also employs cascade diffusion models to enhance the quality of video generation. Inspired by these cascade network designs, we attempt to progressively extract 3D information from a single image to improve the novel view synthesis quality.

\section{Methods}
\label{sec:Methods}
\begin{figure}[t]
\begin{center}
\includegraphics[width=\textwidth]
{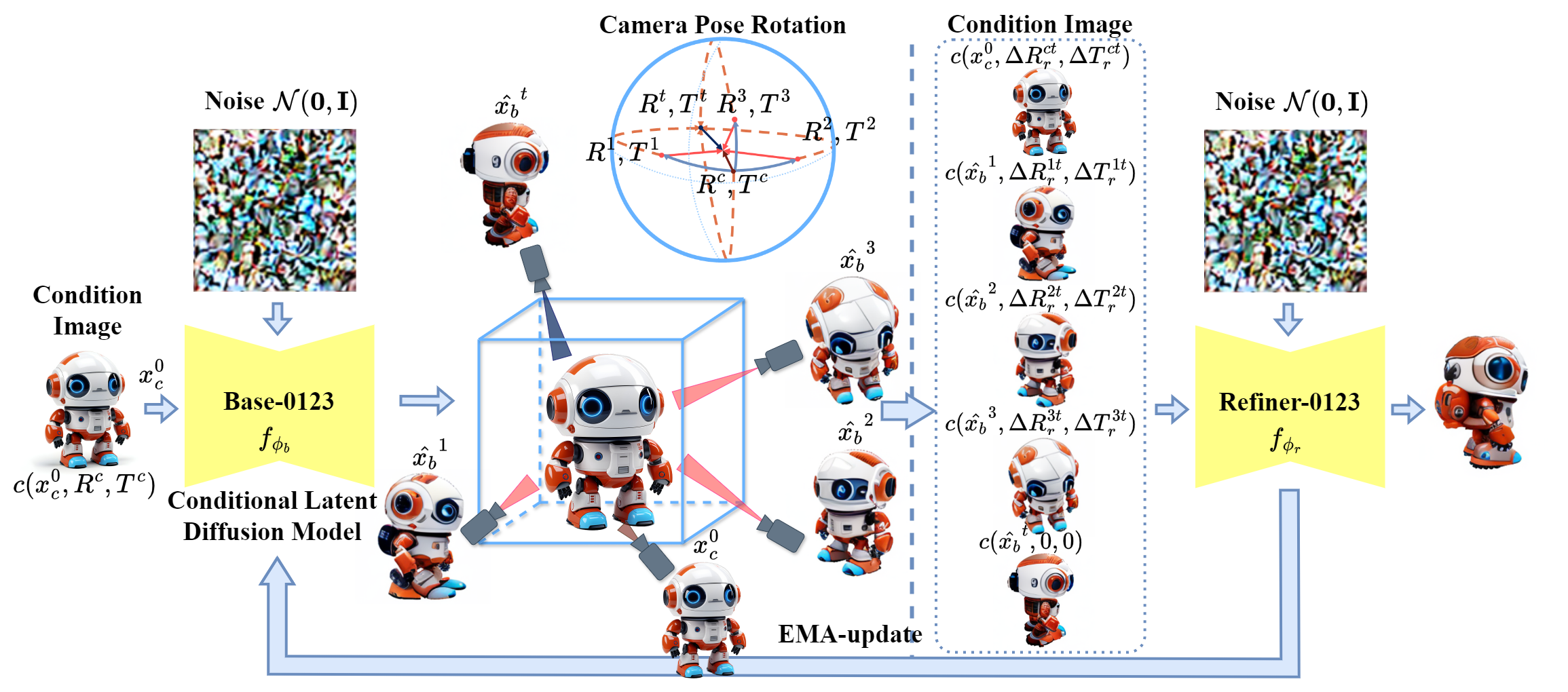}
\end{center}
\caption{The architecture of \name. \name can be divided into two parts. The left part is \stageonename, which takes a set of R and T values as input to generate corresponding multi-view images. These output images are concatenated with the input condition image and its corresponding camera pose, forming a self-prompted input denoted as a set of $c(x_c,\Delta R, \Delta T)$ for the right part \stagetwoname. The corresponding camera pose transition for each condition image to the target image needs to be recalculated as shown in detailed camera pose rotations. After each iteration of training, \stageonename is updated through exponential moving average (EMA) using \stagetwoname.}
\label{fig2_structure}
\end{figure}

We present the \name approach from the following aspects. First, we briefly review the framework of Zero-1-to-3~\cite{liu2023zero1to3} in Sec.~\ref{subsec:Preliminary}. Then, we explain how we construct the cascade structure of Zero-1-to-3 in Sec.~\ref{subsec:structure}. Next, we discuss the design of \stageonename in Sec.~\ref{subsec:basestructure} and \stagetwoname to use self-prompted views in Sec.~\ref{subsec:prompted}. Finally, we describe the self-distillation design in Sec.~\ref{subsec:distillation} and the inference process of the model in Sec.~\ref{subsec:inference}.

\subsection{Preliminary}
\label{subsec:Preliminary}
We first give a brief introduction to diffusion models~\cite{Dhariwal2021diffusion}. The diffusion model's forward process adds the Gaussian noise $\mathcal{N}(\mathbf{0}, \mathbf{I})$, which can be defined as:
\begin{equation}
q\left(x^t \mid x^{t-1}\right)=\mathcal{N}\left(x^t ; \sqrt{\alpha_t} x^{t-1},\left(1-\alpha_t\right) \mathbf{I}\right),
\end{equation}
where $\alpha$ is a scheduling hyper-parameter and $t \in[1,1000]$ denotes the diffusion timestep. $q\left(x^t \mid x^{t-1}\right)$ estimate the probability of $x^t$ using $x^{t-1}$, and $\mathbf{I}$ is the normally distributed variance.

Zero-1-to-3~\cite{liu2023zero1to3} proposes a latent diffusion model to learn the relationship between the source image $x_c$ and target image $x_{t}$ during the denoising process, which can be simply defined as:
\begin{equation}
x_{t}=f_\phi(x_c, \Delta R^{ct}, \Delta T^{ct}).
\label{eq:equition2}
\end{equation}
The Zero-1-to-3 model takes an image $x_c$ and the relative camera pose $(\Delta R^{ct}, \Delta T^{ct})$ as the condition. 
For simplicity, we use $\Delta$ to represent the pose transition, including the relative angle rotation and translation from the condition view $(R^c, T^c)$ to the target view $(R^t, T^t)$, \ie $(\Delta R^{ct}, \Delta T^{ct}) = (R^t, T^t) \ominus (R^c, T^c)$, where $\ominus$ is the pose transition in the corresponding world coordinate and $R$ and $T$ represent the rotation and translation matrix of the camera pose respectively.
Specifically, the pose transition computation of the Objaverse dataset coordinate~\cite{deitke2023objaverse,objaverseXL} can be found in the appendix.

Using a latent diffusion model with an encoder $f_\phi$, a denoiser U-Net $\epsilon_\theta$ and a decoder $\mathcal{D}$, the denoising process can be defined as follows:
\begin{equation}
\begin{aligned}
p\left(x^{t-1} \mid \right.&\left. x^t, c\left(x_c, \Delta R^{ct}, \Delta T^{ct} \right) \right)\\
=\mathcal{N}\left(\right.&\left. x^{t-1} ; \mu_\theta\left(x^t, t, c\left(x_c, \Delta R^{ct}, \Delta T^{ct}\right)\right), \right.\\
&\left. \Sigma_\theta\left(x^t, t, c\left(x_c, \Delta R^{ct}, \Delta T^{ct}\right)\right)\right) .
\end{aligned}
\end{equation}
where $c\left(x_c, \Delta R^{ct}, \Delta T^{ct}\right)$ represents the embedding encoded from the condition image and relative camera pose and the mean distribution $\mu_\theta$ and variance function $\Sigma_\theta$ is modeled by the denoising U-Net.

At the diffusion time step $t$, Zero-1-to-3 encodes the embedding of the input view and relative camera pose as $c\left(x_c, \Delta R^{ct}, \Delta T^{ct}\right)$ and set the objective loss function as:
\begin{equation}
\begin{aligned}
L(\theta)=\min _\theta \mathbb{E}_{z \sim \mathcal{E}(x), t, \epsilon \sim \mathcal{N}(0,1)}
\left\|\epsilon-\epsilon_\theta\left(z_t, t, c\left(x_c, \Delta R^{ct}, \Delta T^{ct}\right)\right)\right\|_2^2,
\end{aligned}
\end{equation}
where $\epsilon$ is the noise prediction corresponding to the distribution.
With the model $\epsilon_\theta$ trained, the inference model $f_{\phi}$ can generate an image by denoising the Gaussian noise conditioned on the embedding of $c\left(x_c, \Delta R^{ct}, \Delta T^{ct}\right)$.

\subsection{\name Framework} 
\label{subsec:structure}
Suffering from the information sparsity of a single
image, it is highly challenging for Zero-1-to-3~\cite{liu2023zero1to3} to generate geometry and appearance consistent novel views with a single source image.
We ease this task with a cascade structure, which progressively moves the camera pose with slight changes.

As shown in Fig.~\ref{fig2_structure}. \name consists of two cascade Zero-1-to-3 models. The first Zero-1-to-3 is referred to as \stageonename, and the second one is referred to as \stagetwoname. 
In terms of the structure, \stageonename is responsible for generating multi-view images. 
These images serve as rough condition inputs, which, along with their corresponding camera poses, are fed into the network of the \stagetwoname model. 
The \stagetwoname is aware of the multi-view input and it will compute the rotation and translation of different viewpoints to the final target view.
Supplying more condition images from different views to \stagetwoname can ease the difficulty of generating novel views.
After that, the training framework incorporates the model parameters of the second \stagetwoname model back into the \stageonename through exponential moving average (EMA).

Compared with Eq.~\ref{eq:equition2}, we can formulate the proposed \name as:
\begin{equation}
\begin{aligned}
x_T=f_{\phi_r}(&(x_c^{0}, f_{\phi_b}(x, \Delta R_b^{c\{1:P\}}, \Delta T_b^{c\{1:P\}})),\Delta R_r^{\{1:P\}t}, \Delta T_r^{\{1:P\}t}),
\end{aligned}
\end{equation}
where $f_{\phi_b}$ and $f_{\phi_r}$ are the \stageonename and \stagetwoname respectively, and the parameters of $f_{\phi_b}$ can be denoted as $\phi_b$ and those  parameters of $f_{\phi_r}$ can be denoted as $\phi_r$. $x_T$ is the final target image. 
$P$ is the number of the prompt views, $\Delta R_r^{it}, \Delta T_r^{it}$ is the prompted views camera pose of \stagetwoname. For easy understanding, we include Table~\ref{table: symbol} to list some symbols mentioned in the paper along with their meanings.

\begin{table}[htbp]
\footnotesize
\centering
\caption{Table of mentioned symbol representation of this paper.}
\begin{tabular}{m{1.7cm}<{\centering}|m{9.8cm}<{\centering}}
\hline
Symbol & Meaning \\
\hline
$x_{c}$, $x_{t}$  & source image and target image  \\
$\Delta R^{ct}$, $\Delta T^{ct}$ & rotation and translation from the condition pose to the target pose \\
$\Delta R_b, \Delta T_b$ & rotation and translation condition to Base-0123\\
$\Delta R_r, \Delta T_r$ & rotation and translation condition to Refiner-0123\\
$f_{\phi_b}$, $f_{\phi_r}$ &Base-0123 model and Refiner-0123 model \\
$\ominus$ & pose transitions in the world coordinate \\
$\hat{x}_{t}^1,...,\hat{x}_{t}^P$ & \small{self-prompted views} \\
\hline
\end{tabular}
\label{table: symbol}
\end{table}

\subsection{\stageonename Framework} 
\label{subsec:basestructure}
In particular, at the beginning of the pre-training process, a set of input images denoted as $x_c^{0}$ is provided to \stageonename.
Then we sample some rotation and translation matrices from the nearby viewpoints of the input view, which we denote as $\{(R^{i}, T^{i})| i \in [1,P] \}$, where $P$ is the total number of prompt nearby views. 
We set the poses of $P$ viewpoints as constant ones to avoid the gap between the training and inference phases. Constant poses can also save computation resources due to that these poses can be reused across different targets.

Our ablation experiments have also demonstrated that generating prompt images with small angles produces good performance. Detailed view setting and hyper-parameter selection can be found in the implementation details of Sec.~\ref{subsec:implement}.

All the $P$ prompted viewpoints attended with the target view pose $(R_t, T_t)$ are concatenated with the same input embedding as $c\left(x_c^{0}, \Delta R_b^{ci}, \Delta T_b^{ci}\right)$, $i \in [1,P]$.
Specifically, we calculate:
\begin{equation}
(\Delta R_b^{ci}, \Delta T_b^{ci}) = (R^{i}, T^{i}) \ominus (R^{c}, T^{c}) \; i \in [1,P],
\end{equation}
where $\ominus$ are the pose transitions in the corresponding world coordinate. In this stage, all these poses are drawn as target views.
They are fed into the first \stageonename $f_{\phi}$ in parallel.

Compared with Eq.~\ref{eq:equition2}, we can formulate the \stageonename framework as:
\begin{equation}
\begin{aligned}
\{\hat{x}_{t}^1,\hat{x}_{t}^2,...,\hat{x}_{t}^P\} =& \{ f_{\phi_b}(x_c^{0}, \Delta R_b^{c1}, \Delta T_b^{c1}), \\
&f_{\phi_b}(x_c^{0}, \Delta R_b^{c2}, \Delta T_b^{c2}), \; ... \\
&f_{\phi_b}(x_c^{0}, \Delta R_b^{cP}, \Delta T_b^{cP}) \},
\end{aligned}
\end{equation}
where $\{\hat{x}_{t}^1,\hat{x}_{t}^2,...,\hat{x}_{t}^P\}$ are exactly self-prompted views.

\subsection{\stagetwoname Framework} 
\label{subsec:prompted}
After obtaining the multi-view images generated by the first-stage \stageonename and their corresponding camera poses, we proceed to compute the camera pose rotation changes for each of these images with respect to the final target image. Specifically, we calculate the relative pose transition from self-prompted views to the target views as:
\begin{equation}
(\Delta R_r^{it}, \Delta T_r^{it}) = (R^{t}, T^{t})  \ominus (R^{i}, T^{i}) \; i \in [1,P],
\end{equation}
where $\ominus$ are the pose transitions in the corresponding world coordinate, and $(R_r, T_r)$ is the camera pose of the target image. Because the camera rotation and translation from the target view to itself are zero and the rotation of the input view remains the same, \ie $(\Delta R_r^{tt}, \Delta T_r^{tt}) = (0,0)$, $(\Delta R_r^{ct}, \Delta T_r^{ct}) = (\Delta R_c^{ct}, \Delta T_c^{ct})$
Then as shown in Fig.~\ref{fig2_structure}. The second \stagetwoname takes the input images and self-prompted nearby views $\{\hat{x}_{t}^1,\hat{x}_{t}^2,...,\hat{x}_{t}^P\}$ as input.
Compared with Eq.~\ref{eq:equition2}, we can formulate the \stagetwoname framework as:
\begin{equation}
\begin{aligned}
x_T =& f_{\phi_r}((x_c^0,\Delta R_r^{ct}, \Delta T_r^{ct}), 
(\hat{x}_{t}^1,\Delta R_r^{1t}, \Delta T_r^{1t}),
... \\
&(\hat{x}_{t}^P,\Delta R_r^{Pt}, \Delta T_r^{Pt}),
(\hat{x}_{t}^{t},\Delta R_r^{tt}, \Delta T_r^{tt})).
\end{aligned}
\end{equation}
Reviewing that the conditional denoising autoencoder can control the synthesis
process through inputs context $y$ such as text, semantic maps or images~\cite{rombach2022stablediffusion}, the latent diffusion models use attention-based models. The context cross-attention can be formulated as:
\begin{equation}
\operatorname{Attention}(Q, K, V)=\operatorname{softmax}\left(\frac{Q K^T}{\sqrt{d}}\right) \cdot V.
\end{equation}
\begin{equation}
Q=W_Q^{(i)} \cdot \varphi_i\left(z_t\right), K=W_K^{(i)} \cdot \tau_{\phi_r}(y), V=W_V^{(i)} \cdot \tau_{\phi_r}(y).
\end{equation}
Here, $\varphi_i\left(z_t\right) \in \mathbb{R}^{N \times d_\epsilon^i}$ denotes a (flattened) intermediate representation of the UNet implementing $\epsilon_{\phi_r}$ and $W_V^{(i)} \in \mathbb{R}^{d \times d_\epsilon^i}, W_Q^{(i)} \in \mathbb{R}^{d \times d_\tau}$ and $ W_K^{(i)} \in \mathbb{R}^{d \times d_\tau}$ are learnable projection matrices.
All prompt views are concatenated through the token dimension so that all these conditions are set as the context of the input $\tau_{\phi_r}(y)$, all conditional embeddings can calculate cross attention to the flattened intermediate representation $\epsilon_{\phi_r}$.

Then, the objective loss function of the \stagetwoname model can be set as: 
\begin{equation}
\begin{aligned}
\min _\theta \mathbb{E}_{z \sim \mathcal{E}(x), t, \epsilon \sim \mathcal{N}(0,1)}\left\|\epsilon- \right.
\left. \epsilon_{\phi_r}\left(z_r, t, c(x_c^{0},\hat{x}_{t}^{\{1:p\}}, \Delta R_r^{\{1:p\}}, \Delta T_r^{\{1:p\}})\right)\right\|_2^2 .
\end{aligned}
\end{equation}
After the model $\epsilon_{\phi_r}$ is trained, we can get the \stagetwoname model $f_{\phi_r}$.

\subsection{Self-Distillation Design}
\label{subsec:distillation}
Considering that the \stagetwoname network $f_{\phi_r}$ is trained via back-propagation to minimize the denoising loss. Inspired by the methods~\cite{XinleiChen2020ImprovedBW,he2020momentum,caron2021emerging,chen2022sdae} in the self-supervised learning field. Our ablation experiments demonstrated that even if only accepted a single image, \stagetwoname
still improves performance. To further enhance the \stageonename model, the \stageonename network is updated in a momentum update way using exponential moving average (EMA). 
Specifically, we have denoted the parameters of $f_{\phi_b}$ as $\phi_b$ and those of $f_{\phi_r}$ as $\phi_r$ we update $\phi_b$ by
\begin{equation}
    \phi_b \leftarrow \eta\cdot\phi_b+(1-\eta)\cdot\phi_r,
\label{eq:ema}
\end{equation}
where $\eta \in[0,1)$ is a momentum coefficient to control the magnitude decay of updates from the \stagetwoname to the \stageonename. 

\subsection{Inference}
\label{subsec:inference}
During the inference stage, given an input image, two stages of the DDIM (Denoising Diffusion Probabilistic Model)~\cite{Jiaming2021ddim} schedule are performed. The input image passes through both the \stageonename and \stagetwoname simultaneously.

These views along with their corresponding camera pose rotation and translation differences, are concatenated and input into the \stagetwoname. After computing cross-attention with the context and inputs, another DDIM sampling is performed, generating the final target image.

For the generation from a single image to 3D, the incurred cost is quite similar to the original Zero-1-to-3.
\stagetwoname leverages these condition images to calculate the SDS loss without repeatedly invoking the denoising process of \stageonename.
The process is similar to novel view synthesis, but the \stageonename only needs to be processed once. All the self-prompted nearby views along with the input views are used as condition images and fed into the \stagetwoname network for computing Score Distillation Sampling (SDS)~\cite{poole2022dreamfusion} loss.

\section{Experiments}
\label{sec:Experiments}
\begin{figure}[t]
\begin{center}
\includegraphics[width=\textwidth]
{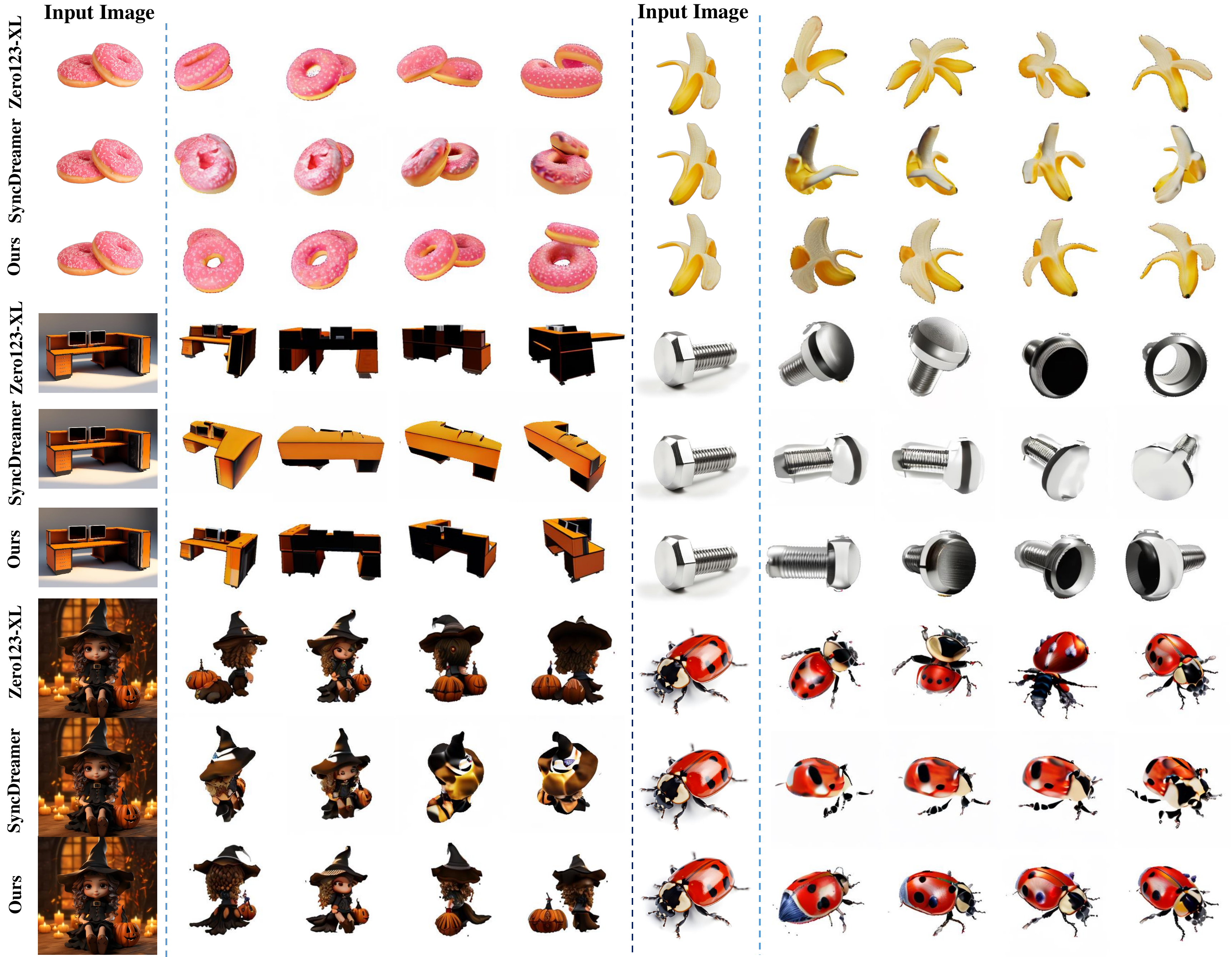}
\end{center}
\caption{Novel view synthesis compared with Zero123-XL~\cite{objaverseXL}, and SyncDreamer~\cite{liu2023syncdreamer}, where Zero123-XL is Zero-1-to-3 pre-trained on Objaverse-XL datasets~\cite{objaverseXL}, achieving higher performance. We selected some challenging scenes, including stacked objects, parallel objects, and objects with multiple branches. Zero123-XL exhibits good quality in image generation but lacks consistency in these complex scenes. SyncDreamer demonstrates good consistency but struggles to maintain good quality in image generation. Our model, however, maintains both quality and consistency in these scenarios.}
\label{fig: fig3_cmp}
\end{figure}
We assess our model’s performance on novel view synthesis and single-image to 3D reconstruction tasks. We introduce the datasets, implementation details, and metrics from Sec.\ref{subsec:Datasets} to Sec.\ref{subsec:metrics}. And show qualitative results, quantitative results on Sec.\ref{subsec:qualitative}. Ablation studies are introduced in supplementary materials. 

\subsection{Datasets}
\label{subsec:Datasets}
\textbf{Objaverse Dataset} is a large-scale dataset containing 800K+ annotated 3D mesh objects~\cite{ouyang2023chasing}. We use this dataset for training and validation. We render 12 images per object from uniformly distributed viewpoints followed by~\cite{liu2023zero1to3}.

\noindent \textbf{Realfusion15} Realfusion15 is the dataset collected and released by RealFusion~\cite{melas2023realfusion}, consisting of 15 natural images.

\noindent \textbf{Google Scanned Object (GSO)}~\cite{downs2022gso} is a high-quality scanned household items dataset. We adopt the same data split as in SyncDreamer~\cite{liu2023syncdreamer}.

\noindent \textbf{RTMV}~\cite{tremblay2022rtmv} consists of complex scenes, and each scene is composed of 20 random objects. We adopt the same data split as in Zero-1-to-3~\cite{liu2023zero1to3}.

\subsection{Implementation Details}
\label{subsec:implement}
We train \name on the Objaverse dataset~\cite{deitke2023objaverse} which contains about 800k objects. Following Zero-1-to-3, the number of viewpoints rendered is 12. Following the assumption of Zero-1-to-3, we also assume that the azimuth of both the input view and the first target view is $0^\circ$. We train the \name for 200k steps with 8 V100 GPUs using a total batch size of 96. 

Both \stageonename and \stagetwoname load the pre-trained Zero123-XL model initially. In \stageonename, to avoid increasing the pretraining costs, we perform only 25 iterations of DDIM (Differentiable Diffusion Model) for inference on each input view of the object during the pretraining phase. Multiple nearby views share the same input view but concatenate with different camera poses. In \stagetwoname, we choose nearby views of azimuth rotations of $45^\circ$ and $-45^\circ$, a view of elevation rotations of 30 degrees, and the target view generated in the first stage is also set as inputs to \stagetwoname. More details about the selection of nearby views can be found in the ablation studies.

\subsection{Metrics}
\label{subsec:metrics}
We use the following evaluation metrics to quantitatively evaluate the performance of our model. We first report the Peak Signal-to-Noise Ratio (PSNR). Perceptual Loss (LPIPS) measures the perceptual distance between two images by comparing the deep features extracted by deep neural networks given each image as input. Structural Similarity (SSIM) measures the structural similarity between two images considering both color and texture information. The CLIP-score (CLIP) quantifies the average CLIP distance between the rendered image and the reference image, serving as a measure of 3D consistency by assessing appearance similarity across novel views and the reference view. We measured Chamfer Distance (C Dist.) to evaluate
point-by-point shape similarity and volumetric IoU (Vol IoU) to quantify the overlap between the reconstructed and the ground truth shape.

\begin{figure}[t]
\begin{center}
\includegraphics[width=\textwidth]
{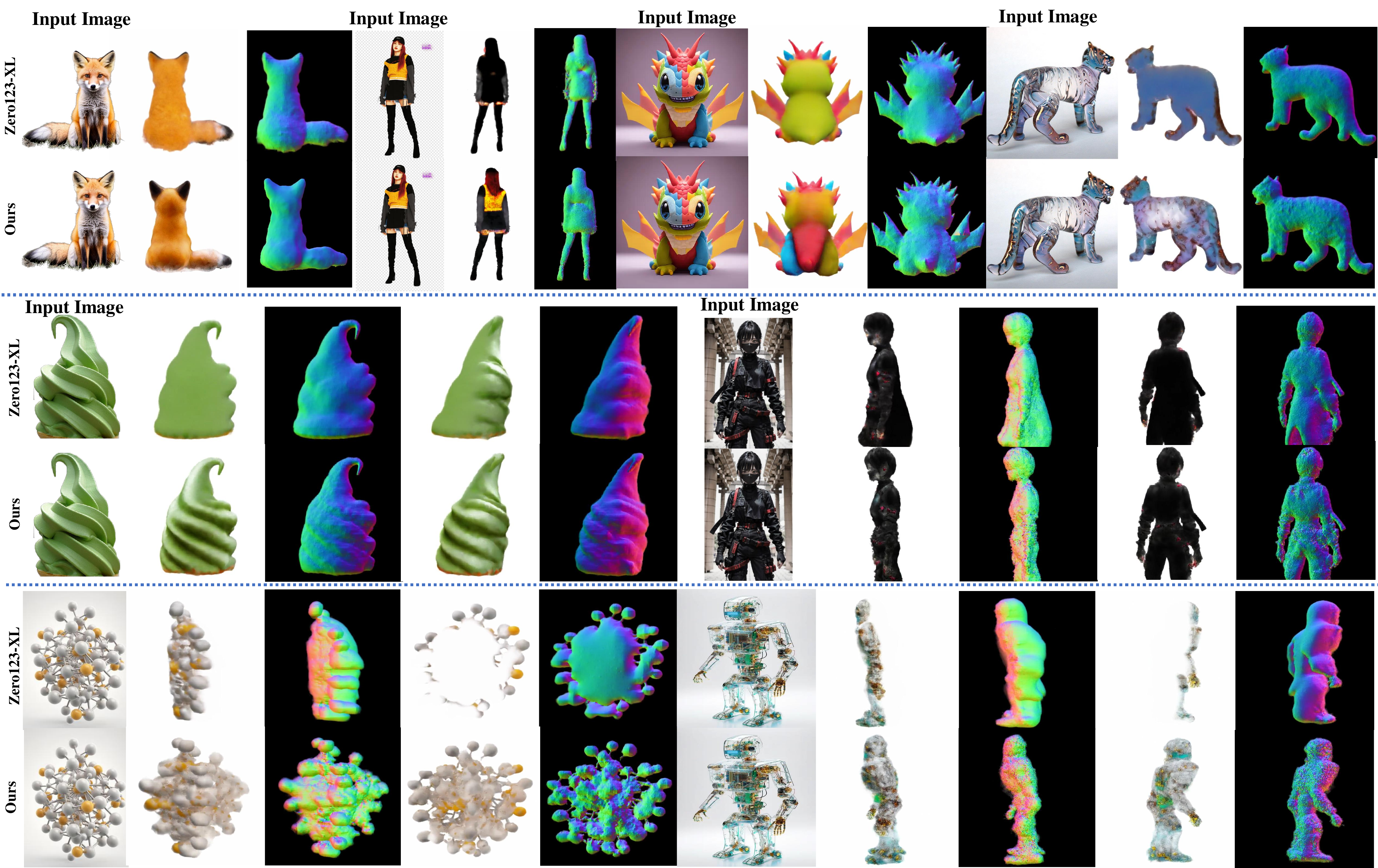}
\end{center}
\noindent\caption{Single image to 3D reconstruction using SDS loss~\cite{poole2022dreamfusion} compared with Zero123-XL. The first two rows illustrate that \name can correct the problem that Zero-1-to-3 sometimes learns inaccurate colors of the backside. The middle two lines describe how \name can rectify structural errors through multi-view self-prompting. The last two lines indicate that \name can address the problem of transparent or high-brightness objects being mistakenly learned as white clouds.}
\label{fig: fig4_sds}
\end{figure}

\subsection{Qualitative Results}
\label{subsec:qualitative}
\textbf{Novel View Synthesis}
We show qualitative results generated by our \name in Fig.~\ref{fig: fig1_consist} and Fig.~\ref{fig: fig3_cmp}. Our model can generalize well to unseen data. We selected various images from real-world scenes or high-quality scenes generated by Stable Diffusion 2.1~\cite{rombach2022stablediffusion} for the experiments on novel view synthesis. These scenes include different kinds of environments and objects. We also tested scenarios involving multiple object stacking (such as stacked donuts) and complex branching structures (such as ladybugs and peeled bananas). The selected example images were deliberately chosen to avoid central symmetry or left-right symmetry, which can pose challenges for generating novel views using conditional latent diffusion models. In the case of these difficult-to-maintain consistency scenes, our \name achieved better consistency compared to Zero123-XL and significantly improved image quality compared to SyncDreamer. Note that Zero123-XL is a model pre-trained on Objaverse-XL~\cite{deitke2023objaverse,objaverseXL} and has superior generation quality.

\noindent\textbf{Single Image to 3D Using SDS Loss}
We show qualitative results of a single image to 3D using Dreamfusion's Score Distillation Sampling (SDS) loss~\cite{poole2022dreamfusion}. All samples are generated by our \name and baseline Zero123-XL in Fig.~\ref{fig: fig4_sds}. Firstly, as shown in the upper two rows of Fig.~\ref{fig: fig4_sds}, after using SDS loss, Zero-1-to-3 tends to learn the backside as the smooth color of the front side. This is due to that the model can only perceive information from the prompt of a single image. As a result, the ears of the fox, the back and hair of Lisa (the woman in yellow), the back of the dinosaur, and the glass tiger are all learned as the same color. 
With the correction provided by self-prompted views, we no longer have to speculate the color of the backside from scratch. Instead, we can progressively predict the backside based on the side views. This allows us to better learn the true color consistency of unseen views.

In addition, the proposal of self-prompted views has enhanced the ability of Zero-1-to-3 to model the shapes and textures of objects as shown in the middle of Fig.~\ref{fig: fig4_sds}. The shapes of the ice cream are well preserved, and even the complex structure of the ninja's back can be modeled. Moreover, Zero-1-to-3, when combined with SDS, has difficulty modeling objects with transparency. Zero-1-to-3 sometimes learns the backside of transparent or high-brightness objects as pure white mist-like clouds. However, the multi-view conditioning information prevents the model from lacking information about other sides, avoiding the learning of a pure white plane for the backside and providing 3D information for objects with transparency.

\begin{table}[t]
\centering
\caption{Quantitative results on Objaverse. We evaluate our method on the test split of Objaverse~\cite{deitke2023objaverse}.}
\begin{tabular}{lcccc}
\hline 
\makebox[0.26\linewidth][l]{Methods} & \makebox[0.11\linewidth][c]{ PSNR$\uparrow$} &  \makebox[0.11\linewidth][c]{SSIM$\uparrow$} & \makebox[0.11\linewidth][c]{LPIPS$\downarrow$} &  \makebox[0.11\linewidth][c]{CLIP$\uparrow$}\\
\hline
Zero123-XL~\cite{liu2023zero1to3,objaverseXL} & 18.68 & 0.883 & 0.189 & 0.758 \\
Magic123~\cite{qian2023magic123} & 18.95 & 0.882 & 0.167 & 0.778 \\
\rowcolor{Lightblue}
Ours & \textbf{21.42} & \textbf{0.911}  & \textbf{0.125} & \textbf{0.802}\\
\hline
\end{tabular}
\end{table}

\begin{table}[t]
\begin{minipage}{0.49\linewidth}
\centering
\caption{Evaluation of novel view synthesis on GSO dataset~\cite{downs2022gso}.}
\begin{tabular}{ccccc}
\hline 
\makebox[0.17\linewidth][l]{\scriptsize{Methods}} & \makebox[0.08\linewidth][c]{\scriptsize{PSNR$\uparrow$}} &  \makebox[0.08\linewidth][c]{\scriptsize{SSIM$\uparrow$}} & \makebox[0.08\linewidth][c]{\scriptsize{LPIPS$\downarrow$}} &  \makebox[0.08\linewidth][c]{\scriptsize{CLIP$\uparrow$}}\\
\hline
\scriptsize{Realfusion}\cite{melas2023realfusion}   & 15.26 & 0.722 & 0.283 & -\\
\scriptsize{Zero-1-to-3}\cite{liu2023zero1to3}   & 18.93 & 0.779 & 0.166 & -\\
\scriptsize{SyncDreamer}\cite{liu2023syncdreamer}   & 20.05 & 0.798 & 0.146 & - \\
\scriptsize{Zero123-XL}\cite{objaverseXL}   & 18.81  & 0.828 & 0.142 & 0.813 \\
\rowcolor{Lightblue}
\tiny{Ours(Zero123-XL)} & \textbf{20.35} & \textbf{0.850} &  \textbf{0.113} & \textbf{0.846}  \\
\rowcolor{Lightblue}
\tiny{Ours(SyncDreamer)} & \textbf{21.10} & \textbf{0.862} & \textbf{0.095} & \textbf{0.944}\\
\hline
\end{tabular}
\label{table: GSO}
\end{minipage}
\begin{minipage}{0.50\linewidth}
\centering
\caption{Evaluation of novel view synthesis on RTMV dataset~\cite{tremblay2022rtmv}.}
\begin{tabular}{ccccc}
\hline 
\makebox[0.17\linewidth][l]{\scriptsize{Methods}} & \makebox[0.08\linewidth][c]{\scriptsize{PSNR$\uparrow$}} &  \makebox[0.08\linewidth][c]{\scriptsize{SSIM$\uparrow$}} & \makebox[0.08\linewidth][c]{\scriptsize{LPIPS$\downarrow$}} &  \makebox[0.08\linewidth][c]{\scriptsize{CLIP$\uparrow$}}\\
\hline
\scriptsize{DietNeRF}~\cite{Jain_2021dietnerf}  &  7.13 & 0.406 & 0.507 & -\\
\scriptsize{Image Variation}~\cite{imagevariation} & 6.56 & 0.442 & 0.564 & -   \\
\scriptsize{SJC-I}~\cite{Wang_2023_sjc_i}  & 7.95 & 0.456 & 0.545 & -  \\
\scriptsize{Zero123-XL}\cite{objaverseXL} & 10.97 & 0.573 &  0.465 & 0.702 \\
\rowcolor{Lightblue}
Ours & \textbf{11.12} & \textbf{0.617} & \textbf{0.441} & \textbf{0.734}  \\
\hline
\end{tabular}
\label{table: RTMV}
\end{minipage}
\end{table}

\begin{table}[t]
\begin{minipage}{0.54\linewidth}
\centering
\caption{Quantitative results on RealFusion15. Evaluation of novel-view synthesis on the RealFusion15 dataset~\cite{melas2023realfusion}.}
\begin{tabular}{lccc}
\hline 
Methods& PSNR$\uparrow$ &  LPIPS$\downarrow$& CLIP$\uparrow$  \\
\hline
RealFusion~\cite{melas2023realfusion} & 20.216 & 0.197 & 0.735\\
Make-it-3D~\cite{tang2023make}  & 20.010 & 0.119 & 0.839\\
Zero-1-to-3~\cite{liu2023zero1to3} & 25.386 & 0.068 & 0.759 \\
Zero123-XL~\cite{objaverseXL} & 25.220 & 0.050 & 0.897\\
Magic123~\cite{qian2023magic123} & 25.637 & 0.062 & 0.747 \\
Consistent-123~\cite{lin2023consistent123} & 25.682 & 0.056 & 0.844 \\
\rowcolor{Lightblue}
Ours & \textbf{26.098} & \textbf{0.043}  & \textbf{0.916} \\
\hline
\label{quantitative_realfusion15}
\end{tabular}
\end{minipage}
\begin{minipage}{0.45\linewidth}
\centering
\caption{Evaluation of single image to 3D reconstruction on GSO dataset~\cite{downs2022gso}.}
\begin{tabular}{ccc}
\hline 
Methods & C Dist.$\downarrow$ & Vol IoU$\uparrow$ \\
\hline
Realfusion~\cite{melas2023realfusion} & 0.0819  & 0.2741\\
Magic123~\cite{qian2023magic123} & 0.0516 & 0.4528 \\
One-2-3-45~\cite{liu2023one} & 0.0629 & 0.4086\\
Point-E~\cite{nichol2022point} & 0.0426 & 0.2875\\
Shap-E~\cite{jun2023shapee} & 0.0436 & 0.3584\\
Zero123~\cite{liu2023zero1to3} & 0.0339 &  0.5035\\
SyncDreamer~\cite{liu2023syncdreamer} & 0.0261 & 0.5421\\
\rowcolor{Lightblue}
Ours &\textbf{0.0207} & \textbf{0.5792} \\
\hline
\end{tabular}
\label{table: CDiou}
\end{minipage}
\end{table}

\subsection{Quantitative Results}
\textbf{Novel view synthesis on Objaverse testset.} Following prior research~\cite{weng2023consistent123}, which randomly picked up 100 objects from the Objaverse testset, 
Since the entire Objaverse test set~\cite{deitke2023objaverse} is quite large, testing all samples would require an excessively long time. Following the approach of Consistent123~\cite{weng2023consistent123}, we randomly selected a subset of samples from the Objaverse test set for testing. These selected samples have not been seen during the training process. Compared with Consistent123~\cite{weng2023consistent123}, we take a larger number of samples for evaluation (Consistent123 only takes 100 samples). Specifically, we randomly selected 200 samples and then randomly selected an input view and a target view following the same setting as Zero-1-to-3. On the Objaverse test set, we measured the reconstruction performance and consistency using metrics of PSNR, SSIM, LPIPS, and CLIP-score (CLIP). 
We utilize the checkpoints from Zero123-XL~\cite{liu2023zero1to3,objaverseXL}.
The index of the selected samples will be made available in the open-source release.

\textbf{Novel view synthesis on Google Scanned Object (GSO).} Following SyncDreamer~\cite{liu2023syncdreamer} and Zero-1-to-3~\cite{liu2023zero1to3}, we adopt the Google Scanned Object~\cite{downs2022gso} dataset as the evaluation dataset. We followed the setting of SyncDreamer~\cite{liu2023syncdreamer} that chose the same 30 objects ranging from daily objects to animals of Google Scanned Object. 
We also apply the proposed cascade mechanism to SyncDreamer~\cite{liu2023syncdreamer}, named as Ours(SyncDreamer). The self-prompted views generated from the first-stage Base-SyncDreamer are fed into the Synchronized Multiview Noise Predictor in the second-stage Refiner-SyncDreamer, as shown in Table~\ref{table: GSO}. We replicate Zero123-XL~\cite{objaverseXL} for fair comparison. We measured the reconstruction performance and consistency using metrics of PSNR, SSIM, LPIPS, and CLIP-score (CLIP). Our cascade method also shows notable improvement on both Zero123-XL and SyncDreamer.

\textbf{Novel view synthesis on RTMV.} In Table~\ref{table: RTMV}, we use the same experimental setup as Zero1-to-3~\cite{liu2023zero1to3} on RTMV~\cite{tremblay2022rtmv} and replicate the performance of the Zero123-XL~\cite{objaverseXL}. We measured the reconstruction performance and consistency using metrics of PSNR, SSIM, LPIPS, and CLIP-score (CLIP).

\textbf{Novel view synthesis on Realfusion15 testset.} Following Magic123~\cite{qian2023magic123} and Consistent123~\cite{lin2023consistent123} We evaluate \name against many related baselines, including RealFusion~\cite{melas2023realfusion}, Make-it-3D~\cite{tang2023make}, Zero-1-to-3~\cite{liu2023zero1to3} and Magic123~\cite{qian2023magic123}, on the RealFusion15 datasets. Like Magic123, we measure the PSNR, LPIPS, and CLIP-scores (CLIP), which assess the reconstruction quality and visual consistency. As shown in Table.~\ref{quantitative_realfusion15}.

\textbf{Single image to 3D reconstruction on GSO.}
In Table~\ref{table: CDiou}, we compare our approach with various other methods following the setting of SyncDreamer~\cite{liu2023syncdreamer}. Both our method and SDS-free methods utilize NeuS~\cite{wang2023neus}, a neural reconstruction method for converting multi-view images into 3D shapes. To achieve faithful reconstruction of 3D mesh that aligns well with ground truth, the generated multi-view images should be
geometrically coherent. \name achieves the
best results in both Chamfer distance (C Dist.) and volumetric IoU (Vol IoU) metrics, demonstrating proficiency in producing multi-view consistent images. 

\section{Limitation and Conclusion}
\label{sec:Conclusion}
\paragraph{Limitation}
\name is based on the pre-trained Zero-1-to-3 model~\cite{liu2023zero1to3}. For cases that are extremely difficult for Zero-1-to-3, \name has limited ability to handle.
With 2D image input, it is challenging to figure out the exact overlap, so even with nearby views, the overlapping parts may still appear to be stuck together. The 3D depth information will be lost and a flat structure will be potentially learned. Additionally, Zero-1-to-3 is sensitive to camera pose elevation. Therefore, \name also struggles with input images that have a high elevation. While our model has already achieved performance and consistency improvement compared to previous models, there are still some potential future directions, \eg enhancing \name by incorporating attention between multiple views as~\cite{shi2023mvdream} and involving multi-modal conditioning information, \eg depth or normal.

\paragraph{Conclusion}
We have witnessed Zero-1-to-3's success in synthesizing novel views from a single image. However, inconsistency still appears as the input information is too limited. We propose a \name network to progressively dig out the multi-view information from the input image by self-prompting nearby views. A series of experimental proofs have demonstrated the effectiveness of this cascade design. We believe the cascade framework can be compatible with more Zero-1-to-3 variants and leave this as an important future work.

\section*{Acknowledgements}
This work was supported in part by the National Natural Science Foundation of China under Grant 62125109, Grant 61931023, Grant 61932022, Grant 62371288, Grant 62320106003, Grant 62301299, Grant T2122024, Grant 62120106007.

%
%
\bibliographystyle{splncs04}
\bibliography{main}

\clearpage

\appendix
\begin{table}[t]
\centering
\caption{Ablation studies on the modules of \name. We evaluate our method on the test split of Objaverse~\cite{deitke2023objaverse}. To conserve resources, we only pre-trained 12K iterations in ablation studies.}
\begin{tabular}{lccc}
\hline 
\makebox[0.45\linewidth][l]{} & \makebox[0.12\linewidth][c]{PSNR$\uparrow$} & \makebox[0.12\linewidth][c]{LPIPS$\downarrow$} &  \makebox[0.12\linewidth][c]{CLIP$\uparrow$}\\
\cline { 2 - 4 } 
Zero123-XL & 18.68 & 0.189 & 0.758\\
\footnotesize{+Cascade two Zero-1-to-3 (frozen)} & 17.67 & 0.189 & 0.778   \\
~\footnotesize{+Prompted random views} & 18.55 & 0.179 & 0.787  \\
~\footnotesize{+Prompted larger views} & 19.04 & 0.172 & 0.783   \\
~\footnotesize{+Prompted more nearby views} & 19.13 & 0.165 & 0.789  \\
\rowcolor{Lightblue}
~\footnotesize{+Prompted nearby views (Ours)} & \textbf{20.04} & \textbf{0.137} & 0.800 \\
\hline
~~~\footnotesize{-coarse target views} & 19.99 & 0.143 & 0.792   \\
~~~\footnotesize{-EMA to \stageonename} & 19.89 & 0.139  & \textbf{0.801}   \\
\hline
\footnotesize{Only single-image prompt for Refiner-0123} & 19.39 & 0.179 & 0.771 \\
\hline
\label{tab:ablation}
\end{tabular}
\end{table}

\begin{figure}[t!]
\begin{center}
\centering
\includegraphics[width=\linewidth]{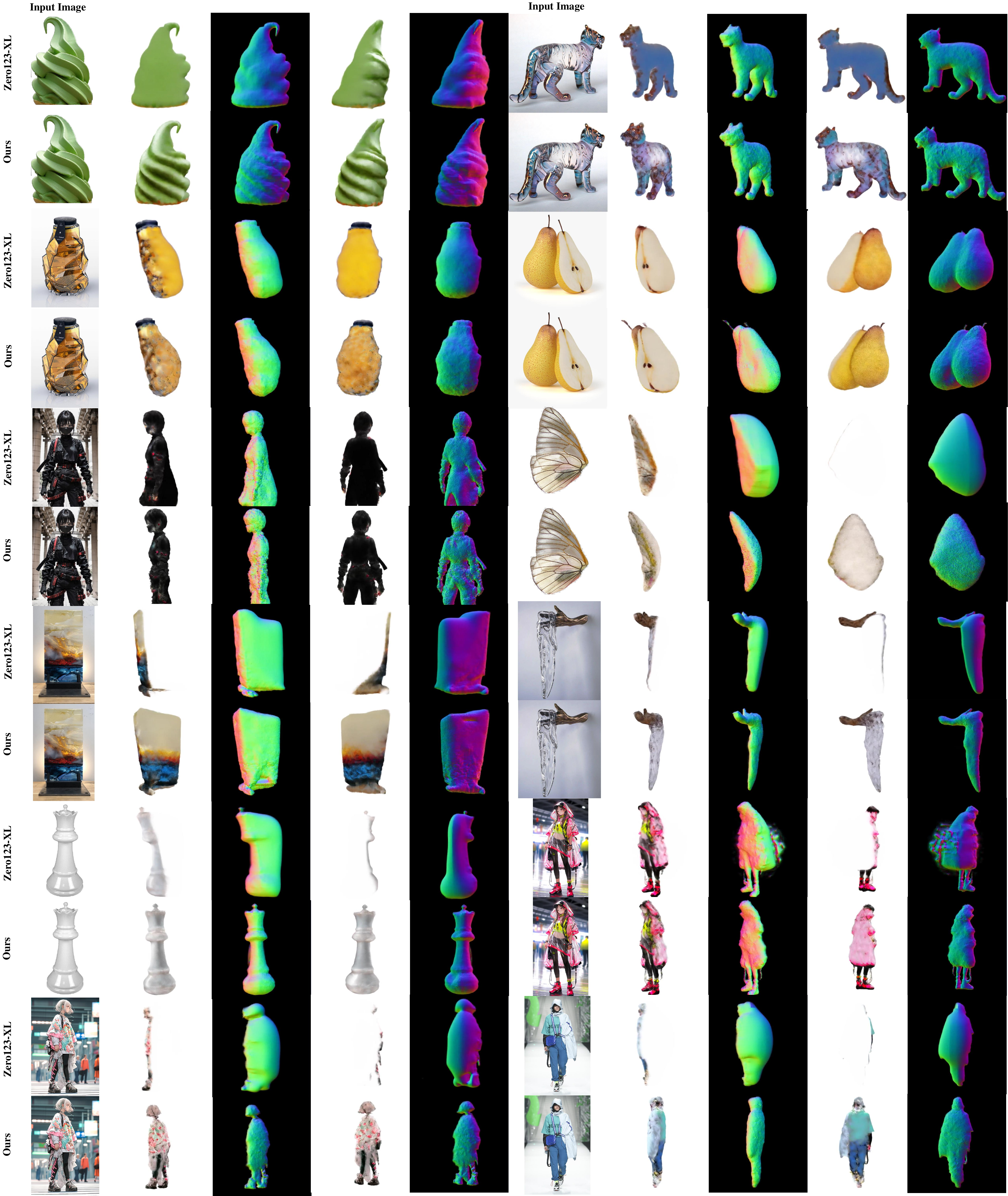}
\captionof{figure}{More qualitative results of single image to 3D reconstruction using SDS loss~\cite{poole2022dreamfusion} compared with Zero123-XL. The first row illustrates that \name can correct the backside through self-prompting. The last row indicates that \name can address the problem of transparent objects being mistakenly learned as white clouds.}
\label{fig:sup3D}
\end{center}
\end{figure}

\begin{figure}[t!]
\begin{center}
\includegraphics[width=\linewidth]{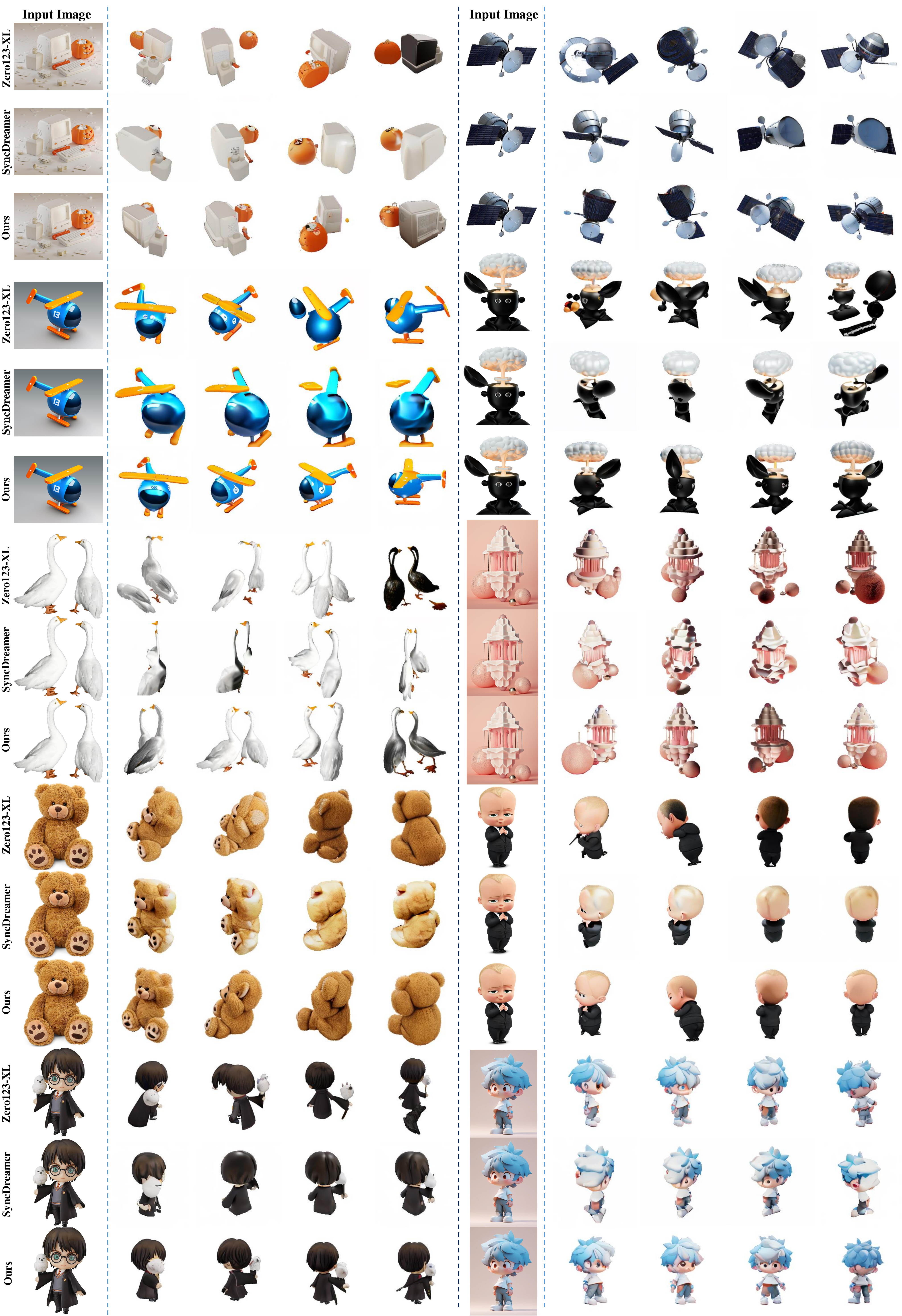}
\end{center}
\captionof{figure}{More qualitative results of novel view synthesis. Rather than adopting the one-to-one generation pipeline as in Zero-1-to-3~\cite{liu2023zero1to3}, \name progressively extracts the 3D information from one single image via self-prompted nearby views. \name shows strong capability on different complex objects compared with Zero123-XL~\cite{liu2023zero1to3,objaverseXL} and SyncDreamer~\cite{liu2023syncdreamer}.}
\label{fig:appendix_nvs}
\end{figure}

\section{Appendix}
\subsection{Ablation Studies}
\label{subsec:ablation}
As shown in Table~\ref{tab:ablation}, in the ablation study, we provide a detailed explanation for the effectiveness of our method, \name, based on the following aspects. Firstly we will study whether nearby views perform better than randomly selected views or larger rotated views. Then we will study the impact of having larger rotated prompted condition views. We will also explore the effects of not using the Exponential Moving Average (EMA) and coarse target views. All the experiments presented below were conducted using the same model, pretraining for 12k iterations on 800K samples from the Objaverse dataset. We experiment with the same testing protocol on the Objaverse testset.
\paragraph{Effects of Self-Prompted Structre} First, we will validate the difference between incorporating prompted views as inputs to the second-stage model compared to using only the fully trained first-stage model. This analysis will measure the gain achieved by the design of \name over the standalone Zero123-XL model. We will then explore various ablations of self-prompted views, 
including completely random prompted views, denoted as "prompted random views"; 
inputting images from surrounding viewpoints, specifically inputting images generated by the \stageonename at $[90^\circ, 180^\circ, 270^\circ]$ angles as the condition images for the second stage, denoted as "prompted larger views"; 
and the effects of having more prompted views on the second-stage model. In practice, we selected the azimuth angles of $[-135^\circ, -90^\circ, -45^\circ, 45^\circ, 90^\circ, 135^\circ]$ and elevation angles of $[-10^\circ, 30^\circ]$. We also conduct ablation on the setting without target views being part of the self-prompted views.
Our model results are highlighted in blue. 

By simply cascading two Zero-1-to-3 models without training and using self-prompted nearby views, we were able to improve the LPIPS and CLIP scores to some extent, but the PSNR decreased. This validates our hypothesis that using multiple viewpoints as conditions can enhance consistency, but simply cascading them may not work. Secondly, using the same number of random views, larger view degree views or even more views with larger view degree views somewhat harm the performance, indicating the need for Zero-1-to-3 to progressively rotate the camera views. Making a large rotation in views at once would lead to large performance degradation. 
\paragraph{Effects of Coarse Target Views} We also include target views as self-prompted views in our experiments. Incorporating target views and generating coarse target views in advance is also beneficial for novel view synthesis.
\paragraph{Effects of EMA Distillations} When the parameters of \stageonename are not updated and kept fixed, the performance will also degrade.
The design of EMA is also shown to have a certain positive effect. Furthermore, As shown in ``Only single-image prompt for Refiner-0123'', even if Refiner-0123 accepts a single image, it still outperforms Zero123-XL. So \stageonename will benefit by inheriting parameters from \stagetwoname.

\subsection{More Qualitative Results of Single Image to 3D}
We show more qualitative results of single image to 3D Dreamfusion's Score Distillation Sampling (SDS) loss~\cite{poole2022dreamfusion}. All samples are generated by our \name and baseline Zero123-XL in Fig.~\ref{fig:sup3D}. 
As shown in the figure, \name is able to correct structural and textural errors, as well as the backside errors of objects. For example, the first row of the figure shows that 
\name fixes the texture and structure errors on the backside of the ice cream. And \name correct the color of the backside of the pears and the structure of the pear pedicle at the same time. 
With the correction provided by self-prompted views, we no longer have to speculate the color of the backside from scratch. Instead, we can progressively predict the backside based on the side views. This allows us to better learn the true color consistency of unseen views.

We also provide qualitative results of multi-view conditioning information to prevent the model from lacking information about transparency, avoiding the learning of a pure white plane for the backside and providing 3D information for objects with transparency. 
For example, the butterfly wings have been learned as solid white blocks instead. 

\subsection{More Qualitative Results of Novel View Synthesis}
We show more qualitative results generated by our \name in Fig.~\ref{fig:appendix_nvs}. Our model can generalize well to unseen data. We selected more various images from real-world scenes or high-quality scenes generated by Stable Diffusion 2.1~\cite{rombach2022stablediffusion} for the experiments on novel view synthesis. These scenes include different kinds of environments and objects. We tested more scenarios involving multiple object stacking (such as two geese, 
a computer and a pumpkin) and complex branching structures (such as a satellite). The selected example images were also deliberately chosen to avoid central symmetry or left-right symmetry, which can pose challenges for generating novel views using conditional latent diffusion models. In the case of these difficult-to-maintain consistency scenes, our \name achieved better consistency compared to Zero123-XL and significantly improved image quality compared to SyncDreamer~\cite{liu2023syncdreamer}. For SyncDreamer, we choose a variety of elevation angles and random seeds to generate images that try our best to show better performance.

\subsection{Cost Analysis}
As shown in Table.~\ref{tab:costs}, we compared the pre-training time per iteration between Zero-1-to-3 and our \name, on the same machine and environment.
It can be observed that our model does not significantly increase the pre-training time. 
As for inference costs, generating one view costs 22 seconds for Zero-1-to-3 and 101 seconds for ours. 
However, generating 100 views as an example, Zero-1-to-3 takes 2203 seconds while our method takes 2356 seconds (6.9\% increased) in total, as generating self-prompted views from the first stage is once-for-all which can be reused. 
Additionally, during the Single image-to-3D stage, there is also only a slight additional inference cost. 

This extra time can be negligible for single image-to-3D tasks that typically take dozens of minutes.

\begin{table}[t!]
\newcommand{\xmark}{\ding{55}}
\centering
\caption{Cost Analysis of \name, compared with Zero-1-to-3~\cite{liu2023zero1to3}
The training time of each iteration.}
\setlength{\tabcolsep}{3pt}
    \begin{tabular}{lccc}
    \toprule
    Methods & $\downarrow$Pretraining Time  & $\downarrow$Inference One View & $\downarrow$Inference 100 Views  \\
    \midrule
    Zero-1-to-3~\cite{liu2023zero1to3}  & 9 seconds & 22 seconds & 2203 seconds\\
    Ours & 15.7 seconds & 101 seconds & 2356 seconds\\
    \bottomrule
    \end{tabular}
\label{tab:costs}
\end{table}

\subsection{Discussion}
The advantage and contribution of Cascade-Zero123 can be summarized that it can progressively dig out the multi-view information but in a reliable range. 
Cascade design is effective and is also adopted in other diffusion models, \textit{e.g.} SDXL~\cite{Podell2023sdxl}. 
More reliable views can lead to better performance in reconstruction models as in PixelNeRF~\cite{yu2021pixelnerf}. We also prove this in generation models via the cascade design by applying the multi-view information on the DM condition.
Compared with directly using Zero 1-to-3 twice, the cascade strategy limits the prompting view in a reliable range. Our ablation study in Table~\ref{tab:ablation} also demonstrates that using unreliable views as conditions can lead to negative effects. It may introduce errors and inconsistencies in the generation process. 
Also, fine-tuning the cascaded model equips it with a capacity to handle multi-view conditions while the original Zero 1-to-3 can only take a single view as the condition. Furthermore, compared with One-2-3-45~\cite{liu2023one2345}, our designed cascade method shows superior performance.

\end{document}